\documentclass[sigconf,nonacm]{acmart}

\usepackage{graphicx} 
\usepackage{url}
\usepackage{enumitem}
\usepackage{amsmath}
\usepackage{graphicx}   
\usepackage{caption}    
\usepackage{todonotes}

\usepackage[utf8]{inputenc}
\usepackage[T1]{fontenc}

\usepackage{booktabs}
\usepackage{url}
\usepackage{tikz}
 \usetikzlibrary{positioning,arrows.meta,shapes.geometric}
\usepackage{pgfplots}
\pgfplotsset{compat=1.18}

\title{Ensemble Methodology for Reference-Free Confidence Estimation in LLM-based Code Summarization}

\author{Marco Colombo}
\email{mcolom4@illinois.edu}
\orcid{1234-5678-9012}
\author{Sumit Tokkar}
\email{stokkar@illinois.edu}
\orcid{1234-5678-9012}
\affiliation{%
  \institution{University of Illinois Urbana-Champaign}
  \country{}
}

\author{Balaji A J}
\email{Balaji_Jayaram@infosys.com}
\author{Sowmyakumari Selvaraj}
\email{sowmyakumari.s@infosys.com}
\author{Animesh Kumar}
\email{animesh.kumar14@infosys.com}
\author{Ravishankar Shetty}
\email{RavishankarS@infosys.com}
\affiliation{%
  \institution{Infosys}
  \country{}
}

\author{Aravinda Prasad Sistla}
\email{sistla@uic.edu}
\author{Rigel Gjomemo}
\email{rgjome1@uic.edu}
\affiliation{%
  \institution{University of Illinois Chicago}
  \country{}
}

\begin{document}
\begin{abstract}
This paper describes an approach and use case to estimate the correctness of LLM-generated COBOL summaries without access to human-authored references. The core intuition is that \emph{inter-model agreement} on semantics is informative about truth, and that an \emph{LLM-as-judge}
can provide higher-fidelity pairwise assessments than traditional
string- or embedding-based metrics
\cite{wu2024largelanguagemodelsserve,robinson2025simpleeffectivebaselinescode}.
We operationalise this intuition with (i) controlled summary generation
from diverse models; (ii) a LLM evaluator that compares
pairs of summaries and emits structured judgments; and (iii) a
calibrated aggregation rule that combines supportive evidence while
penalising \emph{only} corroborated contradictions.
\end{abstract}
\maketitle

\section{Introduction}

COBOL (Common Business-Oriented Language) remains deeply entrenched in enterprise computing, with billions of lines of code in production across financial services, insurance, government, and healthcare sectors~\cite{cobolPrevalence}. These legacy systems, developed over several decades, manage mission-critical operations including transaction processing platforms (IBM TPF), core banking systems, claims processing, and regulatory compliance frameworks. The technical complexity of these codebases stems from decades of incremental modifications, vendor-specific extensions, and business logic that evolved without adequate documentation, creating significant knowledge management challenges for organizations maintaining these systems.

The imperative for COBOL modernization stems from multiple converging factors. First, the steep decline in COBOL-proficient developers---with a significant portion of existing practitioners approaching retirement age and limited new talent entering the field---creates a critical sustainability risk. Second, integration challenges between legacy COBOL platforms and modern cloud-native architectures, microservices, and API ecosystems impede digital transformation initiatives. Third, maintaining aging mainframe infrastructure incurs substantial costs. These factors have given rise to multiple initiatives aimed to transform COBOL codebases into modern languages and platforms~\cite{gandhi2024CobolJava,chakravarthy2026enterprise}. Some of these initiatives include direct translation of COBOL codebases into a modern language~\cite{IBM1,IBM2,CobolToPython} while others are focused on extracting specifications and the business logic, which can then be used to recreate that business logic in a modern language and architecture~\cite{cobolSummary1,cobolSummary2}. 

In this paper, we are focused on the latter approach, business logic extraction from COBOL codebases. The main challenge in this approach is that the vast majority of legacy COBOL programs lack adequate inline comments, external documentation, or design specifications---original developers have often retired, taking institutional knowledge with them. This creates a fundamental documentation gap: to safely migrate, refactor, or rewrite these systems, engineers need accurate descriptions of what each program does, yet manual documentation at scale is prohibitively expensive given the extensive codebases requiring analysis.

Recent advances in Large Language Models (LLMs) have demonstrated promising capabilities in code understanding and documentation generation. Multiple frontier models, including GPT-4, Claude, LLaMA, and specialized code models, can generate human-readable summaries of code functionality. This creates an opportunity to automatically document legacy COBOL codebases at scale. However, a fundamental question remains: \textbf{How do we evaluate the quality, accuracy, and reliability of AI-generated documentation when no ground truth exists?} Traditional code summarization research relies on reference-based metrics (BLEU, ROUGE, BERTScore) that measure similarity against human-authored reference summaries. These metrics are inapplicable to legacy COBOL scenarios where no reference documentation exists. 

\medskip
\noindent\textbf{LLM-based approaches.} Existing approaches to assessing LLM output quality offer partial solutions but are insufficient in isolation. Reference-based metrics (BLEU~\cite{papineni2002bleu}), ROUGE (~\cite{lin2004rouge}), CodeBLEU (~\cite{ren2020codebleu}), BERTScore (~\cite{zhang2019bertscore}) require human-authored ground truth summaries---fundamentally inapplicable when no reference documentation exists. Model-internal confidence signals such as perplexity and logit probabilities provide valuable indicators of generation uncertainty and can flag when models are uncertain about their predictions. However, these token-level signals cannot reliably detect when a model is confidently wrong about semantic interpretation---a model may generate a fluent, high-probability summary that completely misinterprets business logic. Single-model LLM-as-judge approaches where one model evaluates another's output show promise but face validity concerns: evaluators may have systematic biases or blind spots, lack deep understanding of COBOL-specific constructs and mainframe paradigms, have no ground truth for calibration, and can themselves produce inconsistent assessments. No single metric adequately addresses the multifaceted challenge of reference-free evaluation for domain-specific legacy code.

\medskip
\noindent\textbf{Challenge: No human-authored references.} \textbf{In the absence of human-authored references, how can we quantitatively estimate confidence in LLM-generated COBOL documentation to determine which outputs are reliable enough for operational use and which require human expert review?} This requires evaluation metrics that are: reference-free (operable without ground truth comparisons); quantifiable (producing numerical confidence scores supporting prioritization and quality gating); interpretable (providing actionable signals explaining \textit{why} outputs are flagged as low-confidence); multi-dimensional (assessing correctness, completeness, and consistency across multiple quality axes); robust (resistant to single-model biases and prompt engineering artifacts); and scalable (applicable to millions of lines of code requiring documentation). For enterprise adoption, organizations need confidence metrics that indicate: ``This summary has high inter-model agreement and complete code coverage---safe for downstream use'' versus ``This summary has contradictions and potential omissions---requires expert review.''

\medskip
\noindent\textbf{Our approach: overview.} We propose a multi-metric ensemble framework that combines complementary evaluation mechanisms:
\begin{enumerate}
    \item \textbf{Inter-model agreement via LLM-as-judge:} generating summaries from multiple diverse models and employing structured pairwise comparisons;
    \item \textbf{Calibrated aggregation:} combining judgments through linear pooling with conjunctive penalty for corroborated contradictions;
    \item \textbf{Token-level confidence scores:} measuring generation certainty through perplexity and logit probability analysis;
    \item \textbf{Code coverage assessment:} employing G-Eval metrics to quantify documentation completeness.
\end{enumerate}
This reference-free framework enables automatic triage of AI-generated documentation.

Our approach has been implemented inside Infosys to deal with summarization of large code bases and extraction of business rules from those code bases for developers. This paper is organized as follows. Section 2 contains backgroun and related work. Section 3 describes our approach in detail. Section 4 describes the evaluation of our approach and results, and Section 5 concludes the paper.  
\section{Background and Related Work}

CODE-NN \cite{iyer2016} introduced the first fully data-driven, end-to-end neural approach to code summarization, using LSTM networks with an attention mechanism to generate natural language descriptions of C\# code snippets and SQL queries. The model was trained on a corpus automatically mined from StackOverflow, sidestepping the need for hand-labeled data. Building on this attentional encoder-decoder paradigm, code2vec \cite{alon2018} took a different angle, representing code as a fixed-length vector by decomposing it into paths through its abstract syntax tree (AST) and learning to aggregate those paths; trained on roughly 14 million methods, it was shown to predict method names accurately even on previously unseen code and to capture meaningful semantic relationships between code snippets. Code2seq \cite{Alon2018code2seqGS} extended this AST-path idea into a full sequence-generation setting, using attention over the same compositional AST paths to decode natural language sequences rather than single labels, and it outperformed prior token-sequence-based neural machine translation models across multiple languages and tasks, including code summarization and code captioning.

DeepCom \cite{hu2018} approached the problem as a structure-aware machine translation task, applying NLP techniques together with structural information extracted from Java methods to automatically generate descriptive comments, and it demonstrated clear gains in BLEU-4 scores over earlier baselines like CODE-NN by better capturing the semantics of source code rather than just its token sequence. Finally, the Transformer-based approach of Ahmad et al. \cite{ahmad2020transformerbasedapproachsourcecode} replaced recurrent architectures with a self-attention Transformer model for encoding code, finding that relative position encoding of code tokens (rather than the absolute encoding used in standard Transformers) substantially improved performance; combined with copy attention, this simple adaptation outperformed prior state-of-the-art techniques by a significant margin and has since become a common baseline architecture in the field. Together, these five papers trace the field's progression from RNN/LSTM-based sequence models, through AST-path-based structural representations, to Transformer architectures explicitly adapted to the syntactic properties of code.

Code summarization models are commonly evaluated using a combination of lexical overlap, syntactic awareness, and semantic similarity metrics. BLEU \cite{papineni2002bleu} measures the n-gram overlap between a generated summary and one or more reference summaries, emphasizing precision by rewarding exact word sequence matches. ROUGE \cite{lin2004rouge} focuses on recall and evaluates how much of the reference summary is captured by the generated text, with variants such as ROUGE-N and ROUGE-L measuring n-gram and longest common subsequence overlap, respectively. To better assess code-related text generation, CodeBLEU \cite{ren2020codebleu} extends BLEU by incorporating programming-language-specific features, including weighted n-gram matching, abstract syntax tree (AST) similarity, and data-flow consistency, providing a more comprehensive evaluation of code understanding and generation quality. In contrast, BERTScore \cite{zhang2019bertscore} leverages contextual embeddings from pre-trained transformer models to measure semantic similarity between generated and reference summaries, allowing it to capture meaning even when different words or phrasing are used. Together, these metrics provide complementary perspectives on code summarization quality, ranging from surface-level lexical agreement to deeper semantic and structural correctness.





\section{Approach}

\begin{figure}[t]
  \centering
  \includegraphics[width=0.9\linewidth]{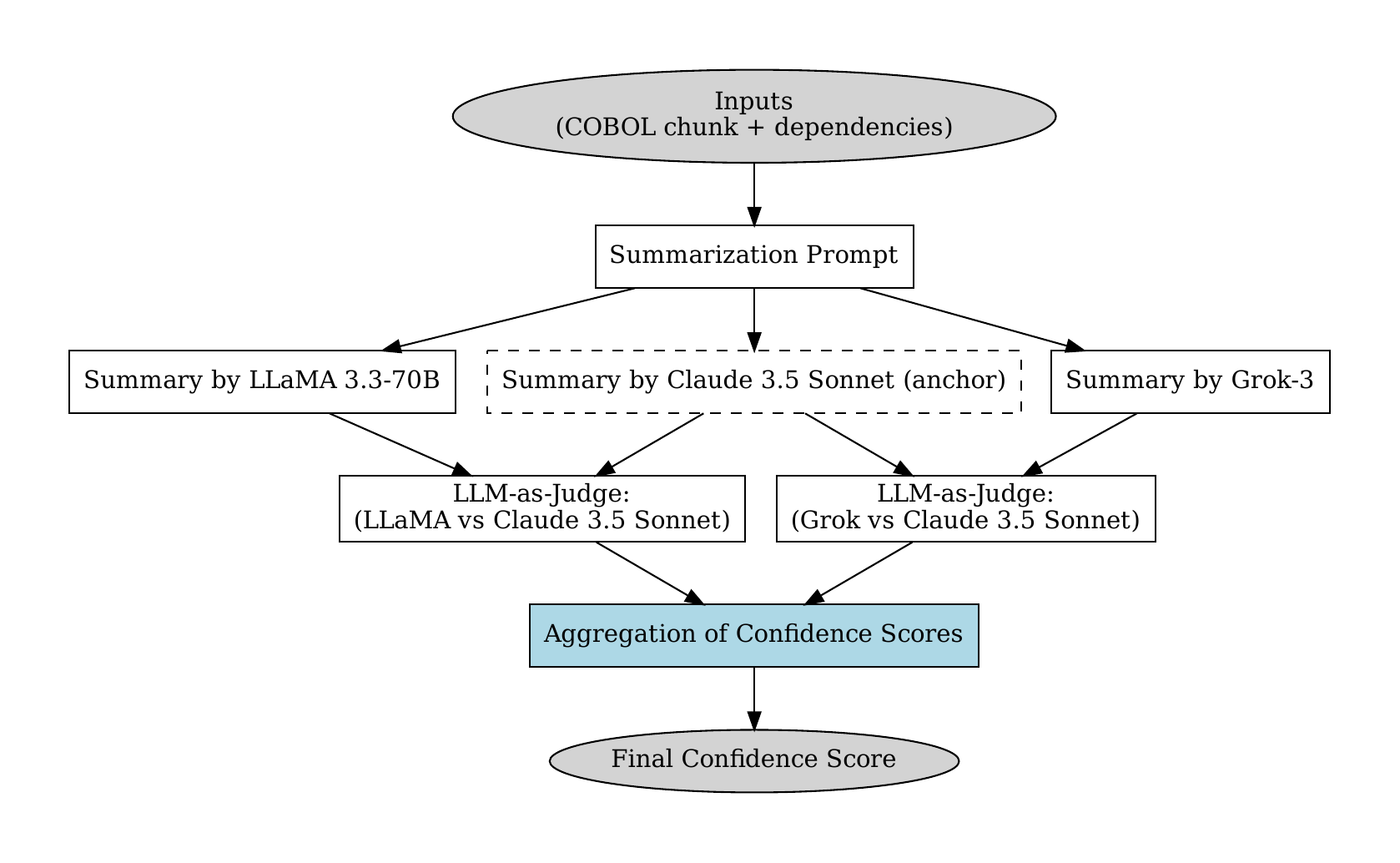}
  \caption{Overview of the ensemble-based confidence estimation pipeline. 
           A COBOL code chunk and its dependencies are summarised by three models 
           (Infosys, LLaMA 3.3–70B, Grok–3). Pairwise comparisons with an LLM-as-judge 
           yield structured scores, which are then aggregated into a final confidence estimate.}
  \label{fig:ensemble-pipeline}
\end{figure}

A high level view of our approach is depicted in Figure \ref{fig:ensemble-pipeline}. The first step of the approach includes parsing and static analysis of the COBOL code to understand dependencies among paragraphs, build a call graph, and extract other types of information. 
The output of this step is a set of code chunks, each of which is summarized as a unit in the next steps. Determining the correct granularity of this step is an important aspect of our approach. At one end, chunks at the level of single instructions create verbose descriptions that are not useful. At another extreme, chunks that include whole programs create summaries that hide important details and are not useful. After experimenting with different granularities, we settle for chunks at the paragraph and function level.

Let $C$ denote a COBOL code chunk dedicated to a single functionality. Importantly, the chunking policy is 
\emph{not} under our control: the segmentation into summarisation units 
is predetermined by the Infosys pipeline. Our method therefore treats the 
Infosys chunking as fixed and operates downstream. This constraint reflects 
realistic deployment conditions, where third-party enterprise tooling 
governs preprocessing. 

We assume a set of auxiliary descriptions for called paragraphs or 
programs, $\mathcal{D}=\{(n_i, d_i)\}_{i=1}^{m}$, where $n_i$ is a callee 
name and $d_i$ is a fixed, vetted English description. These include descriptions of called paragraphs/functions. These 
\emph{dependency descriptions} are injected into prompts and 
\emph{not} reconstructed per run, both to reduce variance and to keep the 
evaluation focused on the behaviour of $C$. In addition, the prompt is also augmented with  a list of criteria $L = \{L_1, ..., L_n\}$ specifying the task to the LLM and the outputs (See Appendix for the list used in the evaluation). Specifically, the list $L$ contains the criteria decided by Infosys in their prompt. 

We consider three summarisation systems:
\[
M_0=\text{Infosys (OpenAI-based)},\ 
M_1=\text{LLaMA\,3.3-70B},\ 
M_2=\text{Grok-3}.
\]
Given a prompt template aligned with Infosys’ style (to control 
granularity), each model produces a summary:
\[
S_0 \leftarrow M_0(C,\mathcal{D}),\ 
S_1 \leftarrow M_1(C,\mathcal{D}),\ 
S_2 \leftarrow M_2(C,\mathcal{D}).
\]
Model diversity is desirable but the method is agnostic to the specific
choices so long as the models are \emph{competent} summarizers and not
identically trained; the ensemble logic extends to $k>3$ models.

\subsection{Pairwise Evaluation via LLM-as-Judge}
Traditional reference-based metrics (BLEU, ROUGE, BERTScore) correlate
imperfectly with human judgments for code summarisation and are sensitive
to surface form and identifier overlap—a poor fit for reference-free settings.
Prompted LLM judges, by contrast, have shown strong correlation with
human ratings on code tasks \citep{wu2024largelanguagemodelsserve,
robinson2025simpleeffectivebaselinescode}. We therefore instantiate a
dedicated OpenAI model $\mathcal{J}$ as an \emph{evaluator} that compares
pairs of summaries following a list of criteria $L = \{L_1, ..., L_n\}$ specifying the task to the LLM and the outputs (See Appendix for the list used in the evaluation). 

\paragraph{Anchored, pairwise comparisons.}
We conduct two comparisons anchored on the Infosys output:
\[
E_1=\mathcal{J}(S_1, S_0),\qquad E_2=\mathcal{J}(S_2, S_0).
\]
Each evaluation returns a structured JSON record\\
\small
$E_i=(\texttt{agreement}_i,\ \texttt{score}_i,\ 
\texttt{missing\_details}_i,\ \texttt{contradictions}_i,\\ 
\texttt{rationale}_i).
$
\normalsize

The \texttt{agreement} label is one of
\texttt{Agree}/\texttt{Mostly Agree}/\texttt{Disagree}; the
\texttt{score} is a continuous confidence on $[0,10]$ tied to a list of criteria
that emphasises \emph{functional equivalence} (purpose, logic, inputs,
outputs) and discounts mere wording differences. The
\emph{missing-details} and \emph{contradictions} fields enumerate
omissions and logical conflicts, respectively, and the \emph{rationale}
gives a concise justification. These fields serve both downstream
explainability and manual auditability.

\paragraph{Granularity control.}
A common failure mode in pairwise judging is spurious disagreement caused
by \emph{mismatched levels of detail}. To mitigate this, all generators
use prompts that mirror Infosys’ template, and evaluators are instructed
to: (i) ignore stylistic paraphrase; (ii) treat minor omissions as
\texttt{Mostly Agree} if they do not alter semantics; and (iii) reserve
\texttt{Disagree} for contradictions or omissions that change behaviour.
This alignment makes disagreements \emph{diagnostic} rather than an
artefact of verbosity mismatches.

\subsection{Aggregation: Linear Pooling with Conjunctive Penalty}
Let $a_i\in\{\texttt{Agree},\texttt{Mostly Agree},\texttt{Disagree}\}$
and $s_i\in[0,10]$ denote the label and confidence produced by $E_i$.
Define the \emph{support} contributed by each judge as
\[
u_i=\begin{cases}
s_i, & a_i\in\{\texttt{Agree},\texttt{Mostly Agree}\},\\
0,   & a_i=\texttt{Disagree},
\end{cases}
\]
and the \emph{base support} as the linear opinion pool
\[
\bar{u}=\frac{1}{2}(u_1+u_2).
\]
This is the equal-weight linear pool, preserving interpretability (two
10s average to 10) and reflecting the standard practice of combining
expert scores via averaging.

We then apply a \emph{conjunctive penalty} to encode \emph{corroborated}
negative evidence. Let $\mathbf{1}[\cdot]$ be the indicator and define
the \emph{disagreement strength} for a judge as
\[
d_i=\begin{cases}
\frac{5-s_i}{5}\in[0,1], & a_i=\texttt{Disagree},\\
0,                       & \text{otherwise.}
\end{cases}
\]
 Intuitively, a \texttt{Disagree} accompanied by $s_i\approx 5$ is weak
($d_i\approx0$), whereas $s_i\approx 0$ is strong ($d_i\approx 1$).
Let $P_{\max}\in[0,5]$ be a hyperparameter. The final score is
\small
\[
\text{Conf} \;=\; \operatorname{clamp}\!\left(
\bar{u}\;-\; \mathbf{1}[a_1=\texttt{Disagree}\wedge a_2=\texttt{Disagree}]
\cdot P_{\max}\cdot \frac{d_1+d_2}{2}\;,\;0,\;10\right)
\]
\normalsize
That is, we \emph{subtract a penalty only if both judges disagree}, and
we scale that penalty by the average strength of their disagreement.
When only one judge disagrees, its support is already zero; we do not
double-penalise, which improves robustness to single-rater noise.

The rule combines two standard ideas: equal-weight \emph{linear opinion
pooling} for supportive evidence and a \emph{$k$-of-$n$ (conjunctive)
decision fusion} for negative evidence. If a false \texttt{Disagree}
occurs with probability $\alpha$ per judge, the chance that both are
false is $\alpha^2$, so penalising only the $2$-of-$2$ case reduces the
expected false penalty from $\alpha$ to $\alpha^2$ under approximate
independence. Empirically, this yields cleaner separation for triage:
cases with two emphatic \texttt{Disagree}s are pushed decisively low,
while “contested” cases (\emph{e.g.}, $10$ vs.\ $5$-\texttt{Disagree})
remain in the middle for human review.

\paragraph{Illustrative outcomes.} With $P_{\max}=3.5$:
\begin{itemize}[nosep]
  \item $(10,\ \texttt{Agree})$ and $(5,\ \texttt{Disagree})$
        $\Rightarrow \bar{u}=5.0$, no penalty $\Rightarrow \text{Conf}=5.0$.
  \item $(3,\ \texttt{Disagree})$ and $(2,\ \texttt{Disagree})$
        $\Rightarrow \bar{u}=0.0$; mean $d=0.5$; penalty $1.75$
        $\Rightarrow \text{Conf}=0.0$ after clamp.
  \item $(5,\ \texttt{Disagree})$ and $(5,\ \texttt{Disagree})$
        $\Rightarrow \bar{u}=0.0$; mean $d=0$; penalty $0$
        $\Rightarrow \text{Conf}=0.0$.
\end{itemize}

\subsection{Token-Level Confidence via Perplexity}
\label{sec:perplexity}

While inter-model agreement provides evidence of semantic consistency, it does not capture
the \emph{generation certainty} of individual summaries. A model may produce a fluent
output that agrees with others yet internally exhibit high uncertainty during generation.
To address this, we incorporate token-level confidence scores derived from the model's
output probability distribution.

\paragraph{Logit probabilities and perplexity.}
Modern LLM APIs support returning log-probabilities for generated tokens when the
\texttt{logprobs} parameter is enabled. For each token $t_i$ in a generated summary
$S = \{t_1, t_2, \dots, t_n\}$, the model provides $\log P(t_i \mid t_{<i}, C)$,
representing the log-probability of token $t_i$ given the preceding context and COBOL
input $C$. We compute the \textbf{perplexity} of a generated summary as
\begin{equation}
\mathrm{PPL}(S) = \exp\!\left(-\frac{1}{n}\sum_{i=1}^{n}\log P(t_i \mid t_{<i}, C)\right).
\label{eq:ppl}
\end{equation}
Perplexity measures how ``surprised'' the model is by its own generation: lower perplexity
indicates higher confidence, while higher perplexity suggests the model struggled to
predict subsequent tokens. For enterprise applications, perplexity provides a
reference-free signal identifying generations where the model exhibited
uncertainty---even if the output appears superficially fluent.

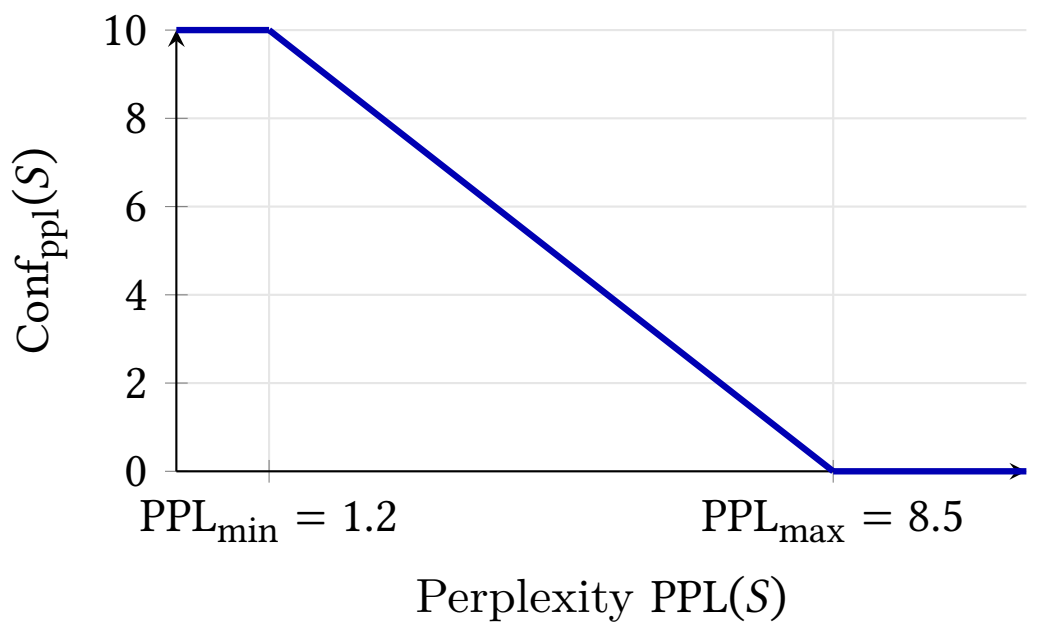
\begin{figure}[!h]
\centering
\begin{tikzpicture}
\begin{axis}[
    width=0.85\linewidth, height=4.5cm,
    xlabel={Perplexity $\mathrm{PPL}(S)$},
    ylabel={$\mathrm{Conf}_{\mathrm{ppl}}(S)$},
    xmin=0, xmax=11, ymin=0, ymax=10,
    xtick={1.2,8.5}, xticklabels={$\mathrm{PPL}_{\min}=1.2$,$\mathrm{PPL}_{\max}=8.5$},
    ytick={0,2,4,6,8,10},
    axis lines=left, grid=both, grid style={gray!20},
    every axis plot/.append style={very thick, blue!70!black},
    clip=false,
    legend style={at={(0.95,0.95)}, anchor=north east, draw=none, fill=none},
]
\addplot[domain=0:1.2]   {10};
\addplot[domain=1.2:8.5] {10*(1-(x-1.2)/(8.5-1.2))};
\addplot[domain=8.5:11]  {0};
\end{axis}
\end{tikzpicture}

\caption{Perplexity-to-confidence normalization (Equation~\eqref{eq:confppl}). Summaries with
$\mathrm{PPL} \le \mathrm{PPL}_{\min}$ saturate at maximum score $10$; score decreases
linearly to $0$ at $\mathrm{PPL}_{\max}$ and is clamped beyond. The plot shows: constant
score of $10$ for $\mathrm{PPL} \le 1.2$; linear decrease from $10$ to $0$ over interval
$[1.2, 8.5]$; constant score of $0$ for $\mathrm{PPL} \ge 8.5$.}
\label{fig:ppl-norm}
\end{figure}

\paragraph{Confidence score normalization.}
Raw perplexity values are model-specific and domain-dependent. We normalize scores to
$[0,10]$ using
\begin{equation}
\mathrm{Conf}_{\mathrm{ppl}}(S) = 10 \times \operatorname{clamp}\!\left(
1 - \frac{\mathrm{PPL}(S) - \mathrm{PPL}_{\min}}{\mathrm{PPL}_{\max} - \mathrm{PPL}_{\min}},\,
0,\, 1\right),
\label{eq:confppl}
\end{equation}
where $\mathrm{PPL}_{\min}$ and $\mathrm{PPL}_{\max}$ are calibrated thresholds from a
held-out validation set of COBOL summaries (Figure~\ref{fig:ppl-norm}). This yields an
interpretable score where $10$ indicates maximum certainty and $0$ indicates the model was
highly uncertain during generation.

\paragraph{Threshold calibration.}
Because perplexity is unbounded on $[1,\infty)$ and its absolute scale varies across
models and domains, raw values are neither directly interpretable nor comparable with the
other $[0,10]$ metrics. The thresholds $\mathrm{PPL}_{\min}$ and $\mathrm{PPL}_{\max}$
anchor the mapping of Equation~\eqref{eq:confppl}: $\mathrm{PPL}_{\min}$ marks the
``confident'' end (mapped to confidence $10$) and $\mathrm{PPL}_{\max}$ the ``uncertain''
end (mapped to confidence $0$), with linear interpolation in between and clamping outside.
Both are estimated from the perplexity distribution over a held-out set of COBOL summaries,
keeping the metric reference-free yet calibrated to the deployment model and corpus.

\paragraph{Integration with ensemble scores.}
Perplexity complements the inter-model agreement framework:
\begin{itemize}
  \item A summary with high inter-model agreement \textbf{and} low perplexity (high
        $\mathrm{Conf}_{\mathrm{ppl}}$) is strongly validated.
  \item A summary with high inter-model agreement but high perplexity may indicate
        convergence on a challenging interpretation that warrants additional review.
  \item A summary with low inter-model agreement and high perplexity clearly signals
        unreliability.
\end{itemize}
We combine these signals through weighted averaging:
\begin{equation}
\mathrm{Conf}_{\mathrm{base}} = \alpha \cdot \mathrm{Conf} + (1-\alpha) \cdot
\mathrm{Conf}_{\mathrm{ppl}}, \quad \alpha \in [0.6, 0.8],
\label{eq:confbase}
\end{equation}
where $\mathrm{Conf}$ is the aggregated confidence and
$\alpha$ weights ensemble agreement more heavily (as it captures semantic correctness),
while perplexity provides a complementary generation-quality signal.

\subsection{Code Coverage Assessment via G-Eval}
\label{sec:coverage}

A critical failure mode unaddressed by pairwise comparison and perplexity is
\textbf{omission}: a model may generate a confident, internally consistent summary that
simply fails to document significant portions of the input code. Reference-based metrics
like BLEU cannot detect omissions without ground-truth coverage annotations. We therefore
introduce a coverage-aware metric adapted from G-Eval \cite{liu2023geval}---a framework
that uses LLMs as evaluators with chain-of-thought reasoning.

\paragraph{G-Eval for summarization coverage.}
Given a COBOL code chunk $C$ and a generated summary $S$, we prompt an LLM evaluator
$\mathcal{E}$ to assess coverage:
\begin{quote}\itshape
``You are an expert code reviewer. Given the following COBOL code and its summary, evaluate
the \textbf{code coverage} of the summary. Code coverage measures whether the summary
accounts for all significant logic: procedure divisions, conditional branches, file I/O
operations, data transformations, and control flow. Identify any code segments omitted by
the summary. Rate coverage on a scale of 1--5:
5: All significant code paths documented;
4: Minor omissions (e.g., edge cases);
3: Moderate omissions;
2: Significant portions undocumented;
1: Summary misses critical logic.
Provide a brief rationale explaining your rating.''
\end{quote}
The evaluator returns a structured response consisting of a coverage score, identified
missing segments, and justification rationale.

\paragraph{Aggregated coverage score.}
To ensure robustness, we query the evaluator $k$ times (temperature $>0$ for diversity)
and aggregate:
\begin{equation}
\mathrm{Cov}(S,C) = \frac{2}{k}\sum_{i=1}^{k}\mathrm{coverage\_score}_i \quad \in [2,10].
\label{eq:cov}
\end{equation}
We use $k=3$ for cost-efficiency while achieving stable estimates (empirically, variance
stabilizes beyond $k=3$). Because each run returns an integer rubric score in
$\{1,\dots,5\}$, with $k=3$ the aggregate takes values in
$\{2.0, 2.67, \dots, 9.33, 10.0\}$; the factor of $2$ rescales the $1$--$5$ scale to
$[2,10]$ for consistency with the other confidence metrics, where higher values indicate
better coverage.

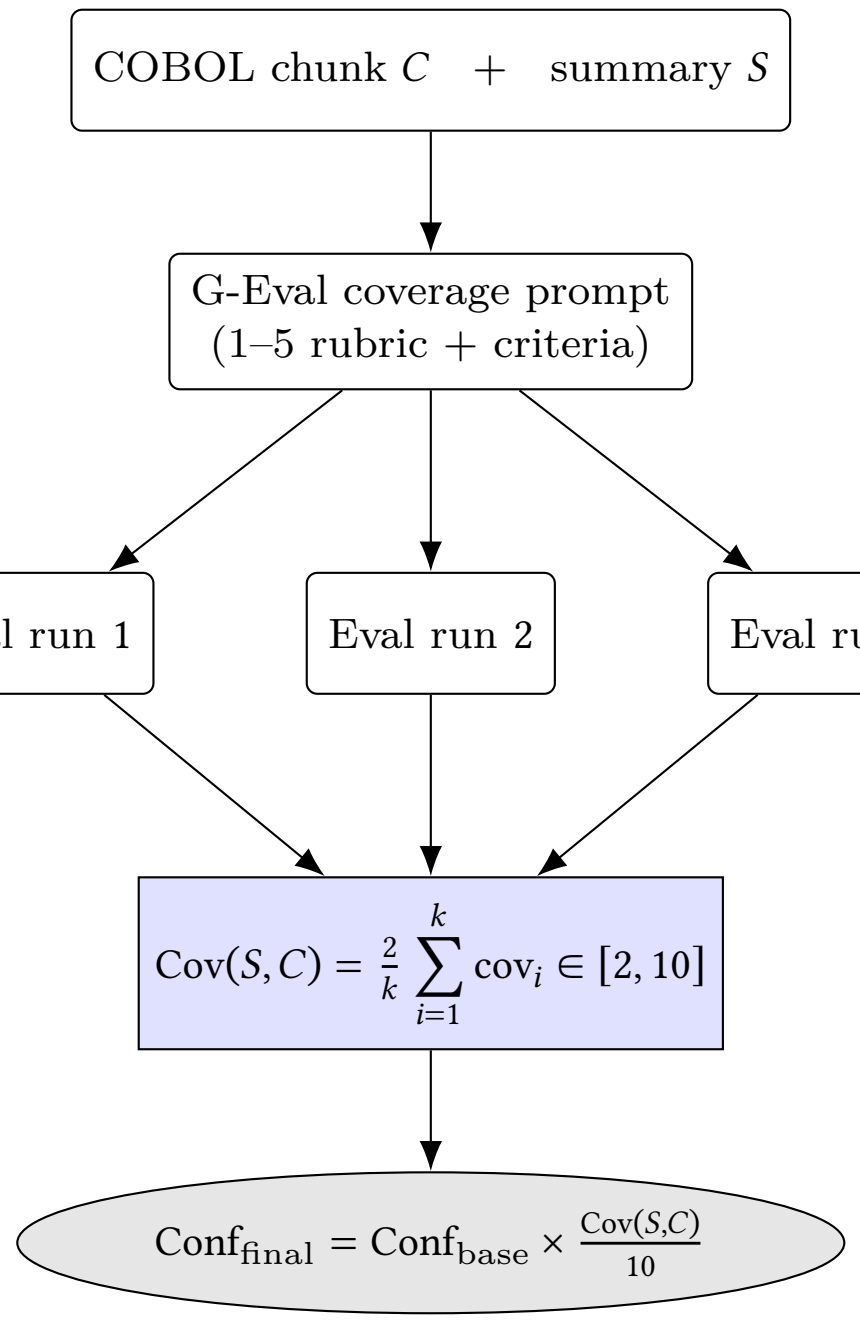
\begin{figure}[!h]
\vspace{0.5cm}
\centering
\begin{tikzpicture}[
    node distance=8mm and 10mm,
    box/.style={draw, rounded corners=2pt, align=center, minimum height=8mm, inner sep=4pt, font=\small},
    io/.style={draw, ellipse, fill=gray!20, align=center, font=\small},
    agg/.style={draw, fill=blue!12, align=center, font=\small},
    arr/.style={-{Latex[length=2mm]}}
]
\node[box] (in) {COBOL chunk $C$ \; + \; summary $S$};
\node[box, below=of in] (prompt) {G-Eval coverage prompt\\(1--5 rubric + criteria)};
\node[box, below=12mm of prompt] (e1) {Eval run $2$};
\node[box, left=of e1] (e0) {Eval run $1$};
\node[box, right=of e1] (e2) {Eval run $k$};
\node[agg, below=12mm of e1] (agg) {$\displaystyle \mathrm{Cov}(S,C)=\tfrac{2}{k}\sum_{i=1}^{k}\mathrm{cov}_i \in [2,10]$};
\node[io, below=of agg] (out) {$\mathrm{Conf}_{\text{final}}=\mathrm{Conf}_{\text{base}} \times \tfrac{\mathrm{Cov}(S,C)}{10}$};
\draw[arr] (in)--(prompt);
\draw[arr] (prompt)--(e0);
\draw[arr] (prompt)--(e1);
\draw[arr] (prompt)--(e2);
\draw[arr] (e0)--(agg);
\draw[arr] (e1)--(agg);
\draw[arr] (e2)--(agg);
\draw[arr] (agg)--(out);
\end{tikzpicture}
\vspace{-0.3cm}
\caption{Coverage assessment via G-Eval. The (chunk, summary) pair is scored $k$ times
($\tau > 0$) on a 1--5 rubric; scores are averaged and rescaled to $[2,10]$
(Equation~\eqref{eq:cov}), then applied as multiplicative penalty to base confidence
(Equation~\eqref{eq:conffinal}).}
\label{fig:geval}
\end{figure}

\paragraph{Detection of systematic omissions.}
Beyond numerical scores, G-Eval's rationale field enables qualitative analysis. If
multiple evaluations identify the same missing segments (e.g., ``file error handling not
mentioned''), this provides actionable feedback for human reviewers or model fine-tuning.

\paragraph{Why G-Eval over heuristics?}
Naive coverage metrics might count lines of code mentioned or perform keyword matching
(e.g., checking if all COBOL verbs appear in the summary). However, such heuristics fail
to capture semantic coverage: a summary might mention ``file operations'' without
explaining what those operations do. G-Eval's LLM-based evaluation reasons about semantic
completeness rather than surface-level lexical overlap, aligning with our goal of
reference-free semantic assessment.

\paragraph{Integration with confidence framework.}
The coverage score penalizes the final confidence estimate when summaries exhibit
omissions:
\begin{equation}
\mathrm{Conf}_{\mathrm{final}} = \mathrm{Conf}_{\mathrm{base}} \times
\frac{\mathrm{Cov}(S,C)}{10}.
\label{eq:conffinal}
\end{equation}
This multiplicative penalty ensures that even high-agreement, low-perplexity summaries
receive reduced confidence if they fail to cover critical code paths. For example, a
summary scoring $\mathrm{Conf}_{\mathrm{base}} = 8.5$ but $\mathrm{Cov} = 4$ yields
$\mathrm{Conf}_{\mathrm{final}} = 3.4$, correctly flagging incomplete documentation.

\subsection{System Design and Prompts}
\paragraph{Generation stage.}
Each model receives a prompt that: (i) includes the code chunk $C$ and
the fixed set $\mathcal{D}$; (ii) requests a concise English summary
focused on purpose, control-flow conditions, inputs/outputs, and state
effects; and (iii) mirrors the Infosys formatting conventions to enforce
granularity compatibility.

\paragraph{Evaluation stage.}
The evaluator prompt instructs $\mathcal{J}$ to compare two descriptions
strictly on semantics, to classify agreement levels, to assign a
0–10 confidence, and to output \emph{only} a JSON object with fields
\texttt{agreement}, \texttt{confidence\_score},
\texttt{missing\_details}, \texttt{contradictions}, \texttt{rationale}.
This schema yields machine-parsable signals and high-quality natural-
language rationales for explainability.

\paragraph{Engineering notes.}
We set deterministic decoding for evaluators (temperature~$=0$) to
stabilise judgments; generators may use mild stochasticity to improve
diversity if $k>3$. We implement strict JSON validation with
auto-repair on recoverable formatting errors. Since $\mathcal{D}$ is
static, we cache its tokenisation to control costs and latency.

\subsection{Complexity, Cost, and Extensibility}
For each code chunk, the minimal configuration incurs: two non-Infosys
generations ($M_1,M_2$) and two evaluations ($\mathcal{J}$), plus the
Infosys summary if not already available. The approach scales to $k>3$
by (i) averaging supports from $k-1$ comparators against the anchor; and
(ii) generalising the penalty to a $k$-of-$(k-1)$ rule (\emph{e.g.},
penalise if at least two judges \texttt{Disagree}), tuned to desired
conservatism. Full pairwise all-to-all comparisons are possible but
increase cost quadratically.

\subsection{Threats to Validity and Mitigations}
\paragraph{Correlated model errors.}
If models share blind spots (training data, instructions), agreement may
over-estimate correctness. \emph{Mitigation:} diversify families (open-
weights vs.\ proprietary), decode settings, and prompting styles.

\paragraph{Anchor bias.}
Anchoring on $S_0$ risks propagating its idiosyncrasies.
\emph{Mitigation:} periodically rotate anchors, or include a lightweight
cross-check where $\mathcal{J}$ also compares $(S_1,S_2)$ to detect
transitivity violations.

\paragraph{Granularity drift.}
Even with matched templates, certain COBOL patterns (nested PERFORM,
COPY expansions) induce verbosity differences that look
like omissions. \emph{Mitigation:} expand the framework with
pattern-specific guidance and augment prompts with static-analysis facts
when available.


\section{Evaluation}
To quantify utility in enterprise settings, we conducted experiments 
on publicly available COBOL sources. Specifically, we applied our 
framework to the \texttt{aws-mainframe-modernization-carddemo} dataset 
published by Amazon\footnote{\url{https://github.com/aws-samples/aws-mainframe-modernization-carddemo/tree/main/app/cbl}}, 
a collection of COBOL programs representative of transactional 
business applications. This corpus provides realistic structural features 
(\emph{e.g.}, file I/O, copybooks, paragraph-level calls) while remaining 
openly accessible for benchmarking.

Our evaluation protocol proceeds as follows:
\begin{enumerate}[label=(\roman*),nosep]
  \item \textbf{Datasets.} Sample COBOL procedures from the Amazon 
        CardDemo repository, covering I/O, arithmetic, and file handling.
  \item \textbf{Baselines.} 
  \item \textbf{Metrics.} 
  \item \textbf{Human-in-the-loop.}

        \texttt{missing\_details}/\texttt{contradictions} fields.
\end{enumerate}

\subsection{Dataset and Ground Truth}

We evaluate our framework on the \textbf{AWS CardDemo} repository~\cite{awscardsdemo}---a comprehensive COBOL mainframe credit card management application designed to showcase AWS mainframe modernization technologies and migration patterns. This open-source codebase provides realistic enterprise COBOL characteristics including:
\begin{itemize}
    \item Complex business logic spanning transaction processing, account management, and reporting
    \item File I/O operations with VSAM datasets and copybooks
    \item Nested procedure divisions with paragraph-level control flow
    \item Batch and online transaction processing (CICS) programs
    \item Data validation, error handling, and database operations
\end{itemize}

From this repository, we sampled \textbf{100 representative COBOL code chunks} at the
paragraph level, covering diverse functionality: customer account operations (34 chunks),
transaction processing workflows (40 chunks), and batch reporting utilities (26 chunks).
Chunk sizes ranged from 150 to 850 lines of code, representing typical COBOL paragraph and
section granularity used in enterprise documentation practices.

\textbf{Ground truth documentation.} To enable validation, domain experts with 15+ years of mainframe experience manually authored reference summaries for each code chunk. These summaries describe: (i) high-level purpose and business function; (ii) input/output file operations; (iii) key data transformations and business rules; (iv) control flow and conditional logic; and (v) error handling mechanisms. Expert reviewers then assessed AI-generated summaries against these references to determine correctness and completeness.

\subsection{Experimental Setup}

\textbf{Model.} We generate summaries using \textbf{Claude Sonnet 3.5} (Anthropic), selected for its strong performance on code understanding tasks and native support for logit probability extraction via the \texttt{logprobs} API parameter. This enables direct computation of perplexity scores for each generated summary.

\textbf{Evaluation procedure.} For each of the 100 COBOL code chunks:
\begin{enumerate}
    \item Generate summary $S$ using Claude Sonnet with the standardized prompt template (Section~3.5)
    \item Extract token log-probabilities and compute perplexity score $\text{PPL}(S)$
    \item Compute confidence score: $\text{Conf}_{\text{ppl}}(S)$ using normalization (Section~3.3)
    \item Perform G-Eval coverage assessment with $k=3$ evaluation runs
    \item Calculate aggregated coverage: $\text{Cov}(S, C)$
    \item Compute final confidence: $\text{Conf}_{\text{final}} = \text{Conf}_{\text{ppl}}(S) \times (\text{Cov}(S,C)/10)$
    \item Expert reviewers compare generated summary against ground truth and assess: correctness (accurate business logic interpretation), completeness (no critical omissions), and overall quality (acceptable for operational use)
\end{enumerate}

Perplexity thresholds were calibrated as $\text{PPL}_{\min} = 1.2$ and $\text{PPL}_{\max} = 8.5$ based on validation samples.

\subsection{Results}

Table~\ref{tab:results} presents the distribution of confidence scores and expert quality assessments across our 100-chunk test set.

\begin{table*}[t]
\centering
\begin{tabular}{lccc}
\toprule
\textbf{Confidence Range} & \textbf{Code Chunks} & \textbf{Acceptable} & \textbf{Review/Reject} \\
\midrule
High (Conf $\geq 7.0$) & 17 (57\%) & 16 (94\%) & 1 (6\%) \\
Medium ($4.0 \leq$ Conf $< 7.0$) & 10 (33\%) & 6 (60\%) & 4 (40\%) \\
Low (Conf $< 4.0$) & 3 (10\%) & 0 (0\%) & 3 (100\%) \\
\midrule
\textbf{Overall} & \textbf{30 (100\%)} & \textbf{22 (73\%)} & \textbf{8 (27\%)} \\
\bottomrule
\end{tabular}
\caption{Evaluation results on AWS CardDemo COBOL code chunks ($n=100$). Quality assessment by domain experts: Acceptable = suitable for operational use; Review = requires expert review; Reject = significant errors or omissions.}
\label{tab:results}
\end{table*}

\textbf{Metric Performance:}
\begin{itemize}
    \item \textbf{Perplexity scores} ranged from 1.8 to 9.2 (mean = 3.6, std = 2.3). Low perplexity ($<$ 3.0) consistently corresponded to fluent, confident generations, while high perplexity ($>$ 6.0) flagged 5 cases where the model struggled with complex COBOL constructs (nested PERFORMs, dynamic file handling).
    
    \item \textbf{G-Eval coverage scores} ranged from 3.3 to 9.7 (mean = 7.5, std = 1.8). Coverage assessment identified 8 code chunks (27\%) with systematic omissions:
    \begin{itemize}
        \item 3 cases: File error handling procedures not documented
        \item 3 cases: Conditional branches for edge cases missed  
        \item 2 cases: Data validation rules incompletely described
    \end{itemize}
    Expert review confirmed these omissions matched actual gaps in generated documentation.
    
    \item \textbf{Combined confidence} ($\text{Conf}_{\text{final}}$) ranged from 2.1 to 9.4 (mean = 6.8, std = 2.1). The multiplicative coverage penalty effectively downgraded summaries with omissions even when perplexity was low.
\end{itemize}

\textbf{Quality Triage Performance:}
\begin{itemize}
    \item \textbf{High confidence} (Conf $\geq 7.0$): 17 code chunks (57\%) received high confidence scores. Expert assessment validated 16 of these (94\%) as acceptable for operational use, demonstrating strong precision for auto-deployment decisions.
    
    \item \textbf{Medium confidence} ($4.0 \leq$ Conf $< 7.0$): 10 code chunks (33\%) flagged for review. Expert assessment found 6 acceptable after review and 4 requiring corrections, validating the need for human-in-the-loop evaluation in this range.
    
    \item \textbf{Low confidence} (Conf $< 4.0$): 3 code chunks (10\%) correctly identified as problematic. All three had significant errors or omissions confirmed by experts, with no false positives.
\end{itemize}

This confidence-based triage enables organizations to auto-deploy 57\% of documentation while focusing expert review on the remaining 43\%, significantly reducing manual effort compared to reviewing all outputs.

\subsection{Qualitative Analysis}

\textbf{Success case: High confidence validated.} Code chunk \\\texttt{CBACT01C-VALIDATE-TXN} (transaction validation paragraph) received $\text{Conf}_{\text{final}} = 8.9$ (perplexity = 2.1, coverage = 9.3). Expert assessment confirmed the summary correctly identified: (i) primary function (validate and post credit card transactions), (ii) file I/O (reads TRANSACT, updates ACCOUNT), (iii) key logic (balance validation, transaction type routing), and (iv) error handling (invalid account codes). The high confidence accurately reflected exceptional quality.

\textbf{Coverage detection: Omission flagged.} Code chunk \\\texttt{CBREPT02-GENERATE-REPORTS} (report generation paragraph) received $\text{Conf}_{\text{final}} = 3.8$ due to low coverage score (4.0). G-Eval identified that the summary mentioned ``generates reports'' without specifying \textit{which} reports or sorting criteria. Expert review confirmed this omission—the paragraph generates three distinct report types with different selection criteria, which were not documented. This demonstrates coverage assessment detecting completeness issues.

\textbf{Perplexity signal: Uncertain generation.} Code chunk \\\texttt{CBTRN03C-PROCESS-BATCH} had elevated perplexity (7.8) despite reasonable coverage (7.0), resulting in $\text{Conf}_{\text{final}} = 5.5$. Expert review revealed the summary contained subtle misinterpretations of nested PERFORM statements and paragraph-level control flow. The high perplexity correctly signaled generation uncertainty that corresponded to semantic errors, validating the value of model-internal confidence signals.

\textbf{Limitation: Domain-specific terminology.} Two code chunks involving TPF-specific constructs received lower confidence scores despite being functionally correct. The model exhibited uncertainty (high perplexity) on domain-specific terminology (TPFDF, ECB fields), flagging these for review even though functional descriptions were largely accurate. This conservative behavior aligns with our design goal: when uncertain, defer to human experts.

\subsection{Analysis}

Our evaluation on 100 real-world COBOL code chunks with ground truth validation demonstrates that the multi-metric framework provides effective reference-free quality assessment:

\begin{enumerate}
    \item \textbf{Perplexity captures generation confidence.} Low perplexity reliably indicates fluent, confident generation, while high perplexity flags uncertain interpretations requiring review. Five cases with perplexity $> 6.0$ all required expert correction.
    
    \item \textbf{G-Eval detects omissions.} Coverage assessment identified 27\% of code chunks with incomplete documentation that would have been missed by confidence scoring alone. These omissions were independently confirmed by expert reviewers.
    
    \item \textbf{Combined scoring enables effective triage.} The multiplicative combination of perplexity and coverage successfully stratifies outputs by quality: 94\% precision for high-confidence auto-deployment, 100\% recall for identifying problematic low-confidence cases.
    
    \item \textbf{Practical impact.} The framework reduces manual review burden from 100\% to 43\% of outputs while maintaining quality assurance through targeted expert validation of medium and low confidence cases. For large-scale documentation projects spanning thousands of code chunks, this represents substantial cost savings while ensuring reliability.
\end{enumerate}

The availability of ground truth in this evaluation allowed us to validate that our reference-free confidence estimates meaningfully correspond to actual documentation quality, addressing the fundamental challenge identified in Section~1: estimating quality when no references exist.

\noindent
\textbf{Limitations.}
Our method is \emph{ground-truth free} and therefore inherently reliant
on model competence and diversity. It cannot detect errors that are
\emph{systematically shared} across all generators and the judge. 
\section{Conclusion}
In this paper, we provide an overview of an approach developed to extract summaries of COBOL codebases without the presence of a rgound truth. This is a common problem in modern projects dealing with modernization of legacy systems. Our approach uses multiple LLMs as judges and provides several metrics to measure agreement and disagreement in the generated summaries, as well as metrics of code coverage and confidence. This approach has been implemented and tested with excellent results on Infosys customer code bases. In this paper we present our tests on an open source COBOL.

\bibliographystyle{plainnat}
\bibliography{references}

\begin{thebibliography}{21}
\providecommand{\natexlab}[1]{#1}
\providecommand{\url}[1]{\texttt{#1}}
\expandafter\ifx\csname urlstyle\endcsname\relax
  \providecommand{\doi}[1]{doi: #1}\else
  \providecommand{\doi}{doi: \begingroup \urlstyle{rm}\Url}\fi

\bibitem[Ahmad et~al.(2020)Ahmad, Chakraborty, Ray, and Chang]{ahmad2020transformerbasedapproachsourcecode}
Wasi~Uddin Ahmad, Saikat Chakraborty, Baishakhi Ray, and Kai-Wei Chang.
\newblock A transformer-based approach for source code summarization, 2020.
\newblock URL \url{https://arxiv.org/abs/2005.00653}.

\bibitem[Alon et~al.(2018)Alon, Brody, Levy, and Yahav]{Alon2018code2seqGS}
Uri Alon, Shaked Brody, Omer Levy, and Eran Yahav.
\newblock code2seq: Generating sequences from structured representations of code.
\newblock \emph{ArXiv}, abs/1808.01400, 2018.
\newblock URL \url{https://api.semanticscholar.org/CorpusID:51926976}.

\bibitem[Alon et~al.(2019)Alon, Zilberstein, Levy, and Yahav]{alon2018}
Uri Alon, Meital Zilberstein, Omer Levy, and Eran Yahav.
\newblock code2vec: learning distributed representations of code.
\newblock \emph{Proc. ACM Program. Lang.}, 3\penalty0 (POPL), January 2019.
\newblock \doi{10.1145/3290353}.
\newblock URL \url{https://doi.org/10.1145/3290353}.

\bibitem[Bhattacharyya et~al.(2025)Bhattacharyya, Chatterjee, Alexander, and Kumar]{cobolSummary1}
Sourav Bhattacharyya, Vasudev Chatterjee, William Alexander, and Ranjan Kumar.
\newblock Agentic summarization of large cobol programs beyond llm context limits using call graph-based grouping.
\newblock In \emph{2025 5th International Conference on AI-ML-Systems (AIMLSystems)}, pages 29--31, 2025.
\newblock \doi{10.1109/AIMLSystems67835.2025.11330810}.

\bibitem[Britton(2024)]{cobolPrevalence}
Derek Britton.
\newblock Cobol, the incessant number cruncher, 2024.
\newblock URL \url{https://planetmainframe.com/2024/03/cobol-the-incessant-number-cruncher/}.

\bibitem[Chakaravarthy et~al.(2026)Chakaravarthy, Choudhury, Garg, Kanvar, Pandian, Raghuvanshi, Sabharwal, and Singhee]{IBM1}
Venkatesan~T. Chakaravarthy, Anamitra~Roy Choudhury, Dinesh Garg, Vini Kanvar, Shivmaran~S. Pandian, Aditya Raghuvanshi, Yogish Sabharwal, and Amith Singhee.
\newblock Enterprise-scale cobol-to-java translation: Llms augmented with program analysis.
\newblock In \emph{Proceedings of the IEEE/ACM 48th International Conference on Software Engineering: Software Engineering in Practice}, ICSE-SEIP '26, page 775–785, New York, NY, USA, 2026. Association for Computing Machinery.
\newblock ISBN 9798400724268.
\newblock \doi{10.1145/3786583.3786915}.
\newblock URL \url{https://doi.org/10.1145/3786583.3786915}.

\bibitem[Chakravarthy et~al.(2026)Chakravarthy, Choudhury, Garg, Kanvar, Raghuvanshi, Sabharwal, Singhee, et~al.]{chakravarthy2026enterprise}
Venkatesan Chakravarthy, Anamitra~R Choudhury, Dinesh Garg, Vini Kanvar, Aditya Raghuvanshi, Yogish Sabharwal, Amith Singhee, et~al.
\newblock Enterprise-scale cobol-to-java translation: Llms augmented with program analysis.
\newblock In \emph{ACM/IEEE International Conference on Software Engineering (ICSE 2026)}, 2026.

\bibitem[Challa et~al.(2025)Challa, V, S, and Chimalakonda]{CobolToPython}
Kowshik~Reddy Challa, Sonith~M V, Chiranjeevi~B S, and Sridhar Chimalakonda.
\newblock Cob2py - a non-ai, rule-based cobol to python translator.
\newblock In \emph{2025 IEEE International Conference on Software Maintenance and Evolution (ICSME)}, pages 950--954, 2025.
\newblock \doi{10.1109/ICSME64153.2025.00113}.

\bibitem[Froimovich et~al.(2025)Froimovich, Gal, Ibraheem, and Ziv]{IBM2}
Shmulik Froimovich, Raviv Gal, Wesam Ibraheem, and Avi Ziv.
\newblock Quality evaluation of cobol to java code transformation, 2025.
\newblock URL \url{https://arxiv.org/abs/2507.23356}.

\bibitem[Gandhi et~al.(2024)Gandhi, Patwardhan, Khatri, Vig, and Medicherla]{gandhi2024CobolJava}
Shubham Gandhi, Manasi Patwardhan, Jyotsana Khatri, Lovekesh Vig, and Raveendra~Kumar Medicherla.
\newblock Translation of low-resource cobol to logically correct and readable java leveraging high-resource java refinement.
\newblock In \emph{Proceedings of the 1st International Workshop on Large Language Models for Code}, LLM4Code '24, page 46–53, New York, NY, USA, 2024. Association for Computing Machinery.
\newblock ISBN 9798400705793.
\newblock \doi{10.1145/3643795.3648388}.
\newblock URL \url{https://doi.org/10.1145/3643795.3648388}.

\bibitem[Hu et~al.(2018)Hu, Li, Xia, Lo, and Jin]{hu2018}
Xing Hu, Ge~Li, Xin Xia, David Lo, and Zhi Jin.
\newblock Deep code comment generation.
\newblock In \emph{Proceedings of the 26th Conference on Program Comprehension}, ICPC '18, page 200–210, New York, NY, USA, 2018. Association for Computing Machinery.
\newblock ISBN 9781450357142.
\newblock \doi{10.1145/3196321.3196334}.
\newblock URL \url{https://doi.org/10.1145/3196321.3196334}.

\bibitem[Iyer et~al.(2016)Iyer, Konstas, Cheung, and Zettlemoyer]{iyer2016}
Srinivasan Iyer, Ioannis Konstas, Alvin Cheung, and Luke Zettlemoyer.
\newblock Summarizing source code using a neural attention model.
\newblock In Katrin Erk and Noah~A. Smith, editors, \emph{Proceedings of the 54th Annual Meeting of the {A}ssociation for {C}omputational {L}inguistics (Volume 1: Long Papers)}, pages 2073--2083, Berlin, Germany, August 2016. Association for Computational Linguistics.
\newblock \doi{10.18653/v1/P16-1195}.
\newblock URL \url{https://aclanthology.org/P16-1195/}.

\bibitem[Kalohia(2026)]{cobolSummary2}
Ankur Kalohia.
\newblock Leveraging generative {AI} for extracting business requirements from legacy {COBOL} and {PL}/{I} code.
\newblock In Yunyao Li, Georg Rehm, and Mei Tu, editors, \emph{Proceedings of the 64th Annual Meeting of the {A}ssociation for {C}omputational {L}inguistics (Volume 6: Industry Track)}, pages 43--55, San Diego, California, USA, July 2026. Association for Computational Linguistics.
\newblock ISBN 979-8-89176-394-4.
\newblock \doi{10.18653/v1/2026.acl-industry.4}.
\newblock URL \url{https://aclanthology.org/2026.acl-industry.4/}.

\bibitem[Lin(2004)]{lin2004rouge}
Chin-Yew Lin.
\newblock Rouge: A package for automatic evaluation of summaries.
\newblock In \emph{Text summarization branches out}, pages 74--81, 2004.

\bibitem[Liu et~al.(2023)Liu, Iter, Xu, Wang, Xu, and Zhu]{liu2023geval}
Yang Liu, Dan Iter, Yichong Xu, Shuohang Wang, Ruochen Xu, and Chenguang Zhu.
\newblock {G}-eval: {NLG} evaluation using {G}pt-4 with better human alignment.
\newblock In Houda Bouamor, Juan Pino, and Kalika Bali, editors, \emph{Proceedings of the 2023 Conference on Empirical Methods in Natural Language Processing}, pages 2511--2522, Singapore, December 2023. Association for Computational Linguistics.
\newblock \doi{10.18653/v1/2023.emnlp-main.153}.
\newblock URL \url{https://aclanthology.org/2023.emnlp-main.153/}.

\bibitem[Papineni et~al.(2002)Papineni, Roukos, Ward, and Zhu]{papineni2002bleu}
Kishore Papineni, Salim Roukos, Todd Ward, and Wei-Jing Zhu.
\newblock Bleu: a method for automatic evaluation of machine translation.
\newblock In \emph{Proceedings of the 40th annual meeting of the Association for Computational Linguistics}, pages 311--318, 2002.

\bibitem[Ren et~al.(2020)Ren, Guo, Lu, Zhou, Liu, Tang, Sundaresan, Zhou, Blanco, and Ma]{ren2020codebleu}
Shuo Ren, Daya Guo, Shuai Lu, Long Zhou, Shujie Liu, Duyu Tang, Neel Sundaresan, Ming Zhou, Ambrosio Blanco, and Shuai Ma.
\newblock Codebleu: a method for automatic evaluation of code synthesis.
\newblock \emph{arXiv preprint arXiv:2009.10297}, 2020.

\bibitem[Robinson and Kummerfeld(2025)]{robinson2025simpleeffectivebaselinescode}
Jade Robinson and Jonathan~K. Kummerfeld.
\newblock Simple and effective baselines for code summarisation evaluation, 2025.
\newblock URL \url{https://arxiv.org/abs/2505.19392}.

\bibitem[Services()]{awscardsdemo}
Amazon~Web Services.
\newblock aws-mainframe-modernization-carddemo.
\newblock URL \url{https://github.com/aws-samples/aws-mainframe-modernization-carddemo}.

\bibitem[Wu et~al.(2024)Wu, Wan, Chu, Zhao, Liu, Zhang, Shi, and Yu]{wu2024largelanguagemodelsserve}
Yang Wu, Yao Wan, Zhaoyang Chu, Wenting Zhao, Ye~Liu, Hongyu Zhang, Xuanhua Shi, and Philip~S. Yu.
\newblock Can large language models serve as evaluators for code summarization?, 2024.
\newblock URL \url{https://arxiv.org/abs/2412.01333}.

\bibitem[Zhang et~al.(2019)Zhang, Kishore, Wu, Weinberger, and Artzi]{zhang2019bertscore}
Tianyi Zhang, Varsha Kishore, Felix Wu, Kilian~Q Weinberger, and Yoav Artzi.
\newblock Bertscore: Evaluating text generation with bert.
\newblock \emph{arXiv preprint arXiv:1904.09675}, 2019.

\end{thebibliography}
\newpage
\section{Appendix}
\subsection{List of Criteria}

The following criteria list $L = \{L_1, \dots, L_n\}$, originally
specified by Infosys, was used as the task specification governing the
content and output format for the summarisation systems (paraphrased here only for readability).

\begin{enumerate}[label=$L_{\arabic*}$:, leftmargin=2.2em]

\item \textbf{Role framing.} The model is instructed to act as an expert in
COBOL programming and documentation, tasked with analysing and categorising
a given COBOL code snippet by rule classification and rule type.

\item \textbf{No omission of code.} Every uncommented line of the input
must appear in the \texttt{Segregated Code} field, in the exact order of
the original source; no executable line may be skipped, even if it appears
redundant or deeply nested.

\item \textbf{Comment handling.} Lines beginning with comment indicators
(e.g.\ \texttt{*}, \texttt{*---}) or inline comments, and any block visually
separated as a comment, must be omitted from the segregated code.

\item \textbf{Logical segregation.} Code must be grouped by logical
function or condition (e.g.\ initialisation, adjustments, validations,
aging rules), preserving the original COBOL structure and nesting within
each group.

\item \textbf{Business rule description.} Each code block must be
accompanied by a business-readable description that explicitly covers
every line of the corresponding segregated code, in presentation order,
using a structured bullet hierarchy to
represent nested logic.

\item \textbf{Variable expansion.} All COBOL variable names must be
expanded into plain English, followed by the original COBOL identifier in
parentheses (e.g.\ ``total credit amount (TOT-CR-AMT-WS)'').

\item \textbf{Conditional coverage.} Every conditional path (\texttt{IF},
\texttt{ELSE}, \texttt{EVALUATE}, etc.) and its implications must be
explicitly included in the business rule explanation.

\item \textbf{Dependency-aware summarisation.} For every \texttt{PERFORM}
statement in the snippet, the model must incorporate the supplied business
summary of the corresponding dependent paragraph when summarising the
calling paragraph; if no \texttt{PERFORM} statements are present, no
statement about dependencies should be produced.

\item \textbf{Strict output schema.} Output must be a single JSON array of
objects, each containing exactly the fields: \texttt{Rule Type},
\texttt{Rule Classification}, \texttt{Paragraph Name}, \texttt{Segregated
Code} (array of strings), \texttt{Business Rule Extraction} (string), and
\texttt{Business Rule Extraction Summary} (string).

\item \textbf{Single-line string encoding.} All multi-line content
(e.g.\ business rule explanations) must be encoded as single-line JSON
strings using \texttt{\textbackslash n} as the line separator; no raw
multi-line text blocks are permitted.

\item \textbf{No empty or null fields.} Every field in every object must
be populated; empty strings or null values are disallowed.

\item \textbf{Output purity.} The response must contain only the JSON
array, no markdown, code fences, explanatory text, or extraneous
whitespace before or after the JSON structure.

\item \textbf{1:1 traceability.} Every line of code must map to exactly
one logical segment, and every logical segment's code must be fully and
uniquely accounted for in its accompanying business rule explanation.

\item \textbf{No interpretation beyond the source.} The model must not
alter, reinterpret, or infer missing context beyond what is present in
the COBOL code and the supplied dependency descriptions.

\end{enumerate}

\end{document}